\documentclass[sigconf]{acmart}

\AtBeginDocument{%
  \providecommand\BibTeX{{%
    \normalfont B\kern-0.5em{\scshape i\kern-0.25em b}\kern-0.8em\TeX}}}

\copyrightyear{2024}
\acmYear{2024}
\setcopyright{acmlicensed}
\acmConference[SIGIR '24] {Proceedings of the 47th International ACM SIGIR Conference on Research and Development in Information Retrieval}{July 14--18, 2024}{Washington, DC, USA.}
\acmBooktitle{Proceedings of the 47th International ACM SIGIR Conference on Research and Development in Information Retrieval (SIGIR '24), July 14--18, 2024, Washington, DC, USA}
\acmISBN{979-8-4007-0431-4/24/07}
\acmDOI{10.1145/XXXXXX.XXXXXX}

\usepackage{multirow}
\usepackage{multicol}
\usepackage[linesnumbered,ruled,vlined]{algorithm2e}
\usepackage{soul}
\usepackage{pifont}
\usepackage{balance}
\newcommand{\ourwork}{\texttt{D$^2$RCU}\xspace}

\begin{document}

\title{Dynamic Demonstration Retrieval and Cognitive Understanding for Emotional Support Conversation}


\author{Zhe Xu}
\orcid{0009-0000-9669-1966}
\affiliation{%
  \institution{Sun Yat-sen University}
  \city{Shenzhen}
  \state{Guangdong}
  \country{China}}
\email{xuzh226@mail2.sysu.edu.cn}

\author{Daoyuan Chen}
\orcid{0000-0002-8015-2121}
\affiliation{%
  \institution{Alibaba group}
  \city{Hangzhou}
  \state{Zhejiang}
  \country{China}}
\email{daoyuanchen.cdy@alibaba-inc.com}

\author{Jiayi Kuang}
\orcid{0009-0009-5764-1398}
\affiliation{%
  \institution{Sun Yat-sen University}
  \city{Shenzhen}
  \state{Guangdong}
  \country{China}}
\email{kuangjy6@mail2.sysu.edu.cn}

\author{Zihao Yi}
\orcid{0009-0009-8656-6374}
\affiliation{%
  \institution{Sun Yat-sen University}
  \city{Shenzhen}
  \state{Guangdong}
  \country{China}}
\email{yizh6@mail2.sysu.edu.cn}

\author{Yaliang Li}
\orcid{0000-0002-4204-6096}
\affiliation{%
  \institution{Alibaba group}
  \city{Hangzhou}
  \state{Zhejiang}
  \country{China}}
\email{yaliang.li@alibaba-inc.com}

\author{Ying Shen}
\authornote{Ying Shen is the corresponding author.}
\orcid{0000-0002-3220-904X}
\affiliation{%
  \institution{Sun Yat-sen University}
  \city{Shenzhen}
  \state{Guangdong}
  \country{China}}
\affiliation{%
  \institution{Guangdong Provincial Key Laboratory of Fire Science and Intelligent Emergency Technology}
  \city{Guangzhou}
  \state{Guangdong}
  \country{China}}
\affiliation{%
  \institution{Pazhou Lab}
  \city{Guangzhou}
  \state{Guangdong}
  \country{China}}
\email{sheny76@mail.sysu.edu.cn}

\renewcommand{\shortauthors}{Zhe Xu et al.}

\begin{abstract}
Emotional Support Conversation (ESC) systems are pivotal in providing empathetic interactions, aiding users through negative emotional states by understanding and addressing their unique experiences. In this paper, we tackle two key challenges in ESC: enhancing contextually relevant and empathetic response generation through dynamic demonstration retrieval, and advancing cognitive understanding to grasp implicit mental states comprehensively. We introduce Dynamic Demonstration Retrieval and Cognitive-Aspect Situation Understanding (\ourwork), a novel approach that synergizes these elements to improve the quality of support provided in ESCs. By leveraging in-context learning and persona information, we introduce an innovative retrieval mechanism that selects informative and personalized demonstration pairs. We also propose a cognitive understanding module that utilizes four cognitive relationships from the ATOMIC knowledge source to deepen situational awareness of help-seekers' mental states. Our supportive decoder integrates information from diverse knowledge sources, underpinning response generation that is both empathetic and cognitively aware. The effectiveness of \ourwork is demonstrated through extensive automatic and human evaluations, revealing substantial improvements over numerous state-of-the-art models, with up to 13.79\% enhancement in overall performance of ten metrics. Our codes are available for public access to facilitate further research and development.
\end{abstract}



\maketitle

\section{Introduction}

As the field of dialogue systems undergoes continuous evolution, the Emotional Support Conversation (ESC) has garnered heightened attention within the dialogue system community \cite{liu2021towards}. An effective ESC focuses more on the emotions of the interlocutors and help them to solve the negative feelings, which indicates that the model treats the interlocutors as help-seekers. To effectively express empathy, ESC necessitates a comprehensive understanding of the help-seeker's experiences and feelings from the history conversations, aiming to mitigate the interlocutor's negative emotional stress and provide guidance to overcome the stress \cite{zhou2018emotional, rashkin2019towards}. The integration of emotional support capabilities into dialogue systems renders them invaluable across various scenarios, markedly enhancing user experience in realms, such as counseling \cite{shen2022knowledge, lin2023emous}, mental health support \cite{sharma2021towards}, and customer service chats \cite{vassilakopoulou2023developing}.

\begin{figure}[t]
  \centering
  \includegraphics[width=\linewidth]{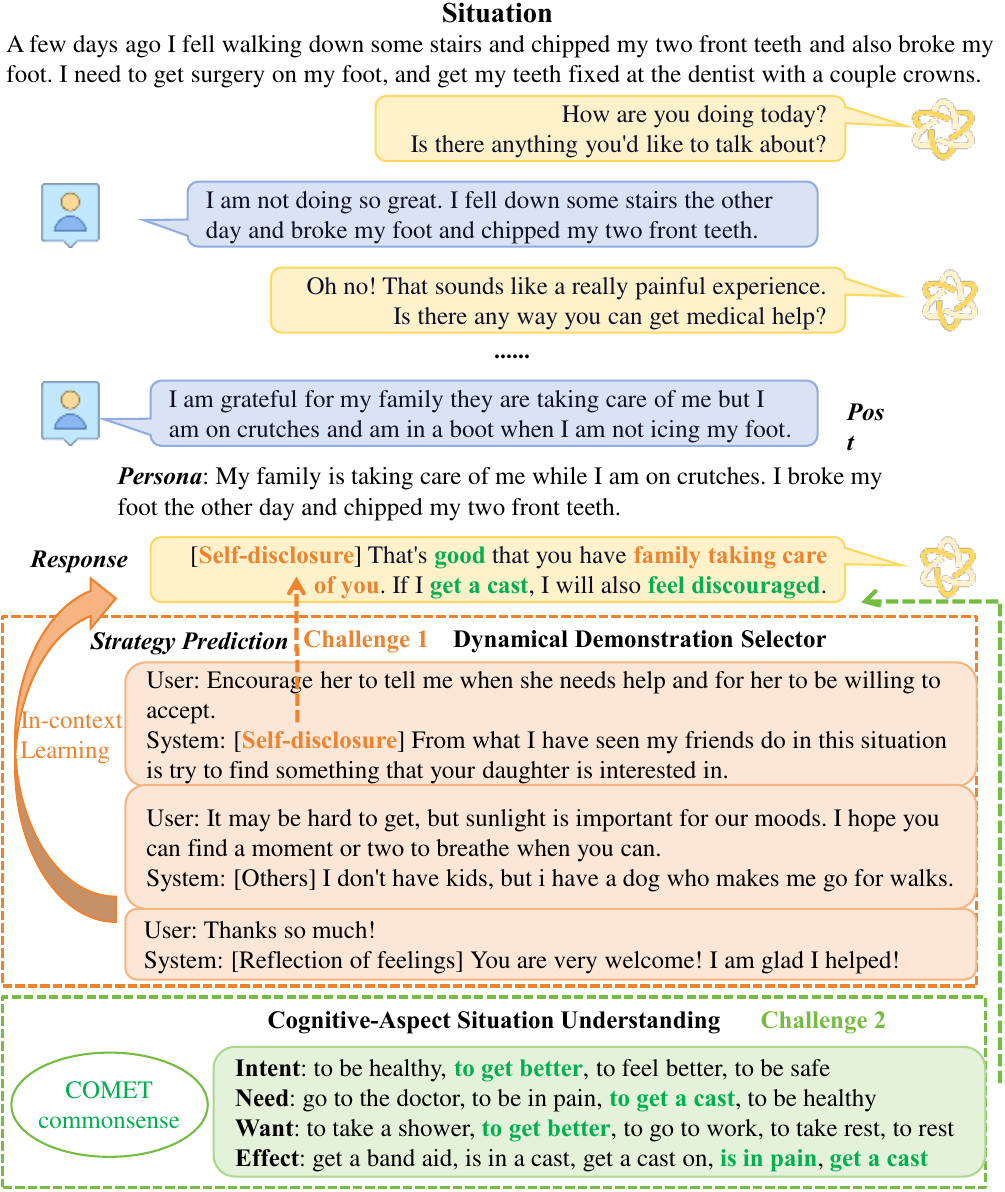}
  \caption{Illustration of an emotional support conversation. For queries posted by help-seeker, knowledge is extracted to support responses through the dynamic demonstration selector and COMET within the proposed \ourwork.}
  \label{Fig:sample}
\end{figure}

As illustrated in Figure \ref{Fig:sample}, in the ESC task, our supporter generates \textit{responses} based on the \textit{post} and \textit{persona} of the help-seeker. A typical ESC scenario is portrayed, constituting a framework with three distinct stages, namely Exploration, Comforting, and Action, as proposed by Hill et al. \cite{hill2009helping}. The ``Situation'' explicates the reason why the help-seeker suffers from the emotional problem from history conversation context. In the ESC procedure, the supporter initiates the interaction by exploring the help-seeker's experiences and situation. MISC \cite{tu2022misc} innovatively integrates COMET, harnessing external commonsense knowledge to enhance the model's adeptness in reasoning about emotions. Cheng et al. \cite{cheng2022improving} employ a specialized detector to obtain the cause of the emotion. Moving to the comforting stage, the supporter expresses understanding and sympathy through various comfort techniques (such as Reflection of Feeling, Affirmation and Reassurance). In the subsequent Action stage, the supporter takes proactive measures to encourage the help-seeker, offering constructive suggestions to address the underlying problem. Cheng et al. \cite{cheng2022improving} devise an A*-like heuristic algorithm to select an optimal response strategy. Addressing the imperative of personalized user responsiveness, PAL \cite{cheng2023pal} construct the user's persona and adopts a strategy-based decoding method.

While contemporary approaches in ESC have shown advancements, they leave two crucial challenges for building and enhancing the capabilities of ESC systems:

\textbf{Challenge 1: how can ESC systems adeptly retrieve and integrate contextual demonstrations to generate responses that are both relevant and empathetic?} 
Traditional end-to-end generative models often struggle with producing contextually appropriate responses, particularly in the realm of open-ended dialogue \cite{sutskever2014sequence, guo2018dialog, li2016diversity}. While efforts have been made to augment these models with retrieval mechanisms to improve response coherence and maintain conversation goals \cite{wu2019sequential, gupta2021controlling}, there remains an uncharted domain in tailoring these retrieval methods to align with the nuanced persona of the user within the specific context of ESC.

\textbf{Challenge 2: how can we advance the cognitive-aspect situation understanding capabilities of ESC systems to grasp the user's implicit mental states comprehensively?} Emotional dialogue is a dual-faceted endeavor encompassing both affective recognition for identifying the user's emotions, and cognitive comprehension for understanding the user's unique circumstances. While recent strides have focused on the affective dimensions of this task \cite{tu2022misc, peng2022control, zhao2023transesc}, the equally pivotal cognitive component is often relegated to the background. Some efforts have been made to incorporate commonsense knowledge into empathic dialogue systems as a means to bolster cognitive empathy \cite{sabour2022cem}, but none have ventured into leveraging cognitive understanding to provide profound empathetic support in ESC systems. 

Addressing these questions is crucial for developing ESC systems capable of navigating the intricate interplay of cognitive insight, thereby crafting responses that genuinely resonate with the user's cognitive and situational needs. To tackle these issues, in this paper, we propose a method called Dynamic Demonstration Retrieval and Cognitive-Aspect Situation Understanding (\ourwork) tailored for emotional support conversations. As depicted in Figure \ref{Fig:sample}, our approach concurrently accomplish \textit{dynamic demonstration retrieval} (\textcolor{orange}{orange-colored region}) and \textit{cognitive understanding} (\textcolor{green}{green-colored region}), guiding response generation through multiple knowledge fusions. Drawing inspiration from In-context Learning \cite{min2022rethinking}, we exploit in-sample knowledge by dynamically retrieving relevant query-passage pairs for each user's post in the training set, aiming to tackle the \textit{Challenge 1}. Moreover, we design a Cognitive-Aspect Situation Understanding algorithm to effectively enhance the comprehension of cognitive aspects involved. By leveraging the knowledge source ATOMIC \cite{sap2019atomic}, we select four cognitive relationships (Effect, Intent, Need, and Want) to gain a deeper understanding of the situations and implicit mental states of help-seekers, employed to address \textit{Challenge 2}. Finally, a supportive decoder that leverages retrieval enhancements and cognitive awareness, integrating information from various knowledge sources to facilitate the decoding process.

Our model demonstrates commendable performance across both automatic and human evaluations. Notably, in the term of automatic evaluation compared with numerous state-of-the-art methods, our model achieves the highest scores in all ten metrics and improves the strongest baseline by an average normalized increase of 13.79\%. Additionally, ablation studies validate the significance of each proposed component, demonstrating that In-context Learning substantially enhances our model's performance.

In summary, our main contributions are as follows:

\begin{itemize}
    \item We present a novel approach for leveraging persona information in dynamic demonstration selection, which yields more informative and personalized candidate pairs.
    \item To explicitly model the cognitive understanding of the user's situation, we introduce a dedicated cognitive understanding module, emphasizing the significance of cognitive awareness, and devise a response generation method based on retrieval enhancement and cognitive understanding.
    \item We conduct extensive experiments and demonstrate that our proposed \ourwork significantly improves the SOTA performance of the ESC task by up to 13.79\%. To facilitate further research, our codes are publicly available at: \url{https://github.com/Bat-Reality/DDRCU}.
\end{itemize}

\section{Related Work}
\subsection{Empathetic Dialogue Systems}

Empathetic dialogue systems are designed to recognize and respond to interlocutor's emotions, creating more meaningful interactions \cite{zhou2018emotional, rashkin2019towards, liu2021towards}. While most current research has focused on the affective aspect of empathy by detecting the interlocutor's emotions \cite{rashkin2019towards, majumder2020mime, li2022knowledge}, there has been a growing recognition of the need to address the cognitive aspect of empathy. This has led to efforts to incorporate cognitive commonsense knowledge, thereby enriching the representations of cognitive empathy \cite{sabour2022cem}. 
Knowledge graph, especially in terms of commonsense and affective concepts, has been explored for empathetic response generation \cite{zhou2022case,chen2019knowledge}. 
Furthermore, the scope of empathy extends beyond the mere generation of responses based on suitable emotions; it entails a profound understanding of the interlocutor's nuanced expressions and situation \cite{majumder2021exemplars}. 
With three communication skills (Exploration, Comforting, and Action), Majumder et al. \cite{majumder2021exemplars} introduce Dense Passage Retrieval (DPR) to extract context-relevant exemplary responses. Despite advancements, fully understanding of the interlocutor's mental states through cognitive aspect of empathy remains challenging. 
Our work extends this line of research by employing a dedicated cognitive understanding module to better grasp and reflect the cognitive states of help-seekers, enhancing the empathetic depth of dialogue systems. 

\subsection{Emotional Support Conversation}

\begin{figure*}[!ht]
    \centering
    \includegraphics[width=0.83\linewidth]{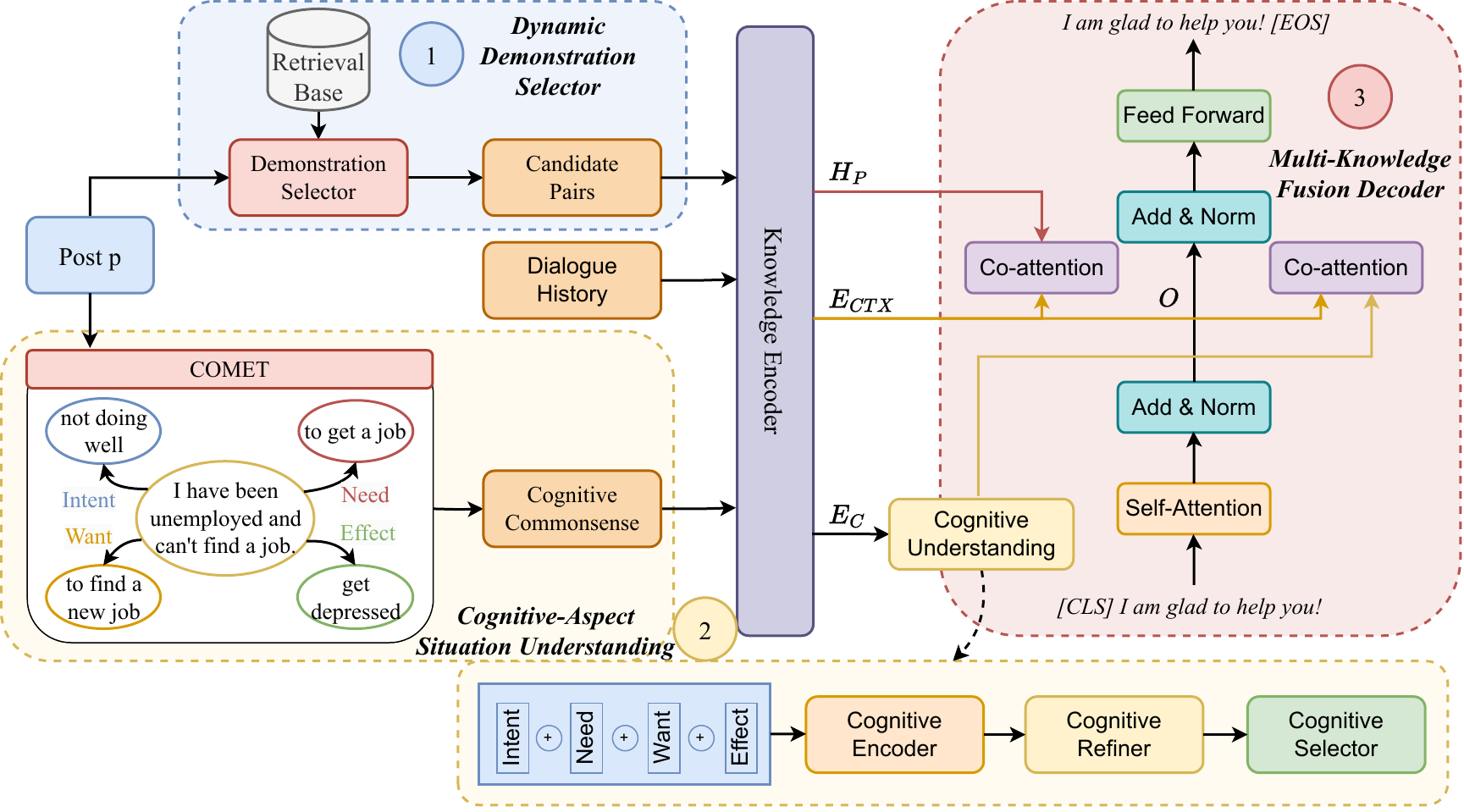}
    \caption{The overall architecture of our proposed D$^2$RCU, comprises three key components: \ding{172}\textit{Dynamic Demonstration Selector}, \ding{173}\textit{Cognitive-Aspect Situation Understanding}, and \ding{174}\textit{Multi-Knowledge Fusion Decoder}. The details of the Cognitive Understanding process are highlighted by the bottom orange box. Given a user's post, we first obtain prior knowledge through dynamic demonstration selection and COMET commonsense extraction, then understand the user's feeling through the cognition of the current situation, and finally generate responses through a knowledge-aware decoder.}
\label{Fig:RECU}
\end{figure*}

Compared with Empathetic Dialogue Systems, ESC goes beyond expressing empathy, aiming to provide guidance with help-seekers through their negative emotional distress \cite{liu2021towards}. Current methods, often based on COMET \cite{bosselut2019comet}, a pre-trained inference model, strive to effectively identify emotional challenges \cite{peng2022control, zhao2023transesc, deng2023knowledge, zhou2023facilitating, li2024dq}. MISC \cite{tu2022misc} integrates commonsense knowledge for fine-grained emotion understanding, and Zhao et al. \cite{zhao2023transesc} explore state transitions (such as, semantic, strategic and emotional) between each conversation turn. Peng et al. \cite{peng2022control} model the relationship between global commonsense causes and local contextual intentions by a hierarchical graph network. 
Several novel strategies, including reinforcement learning, have been adopted to steer emotional discourse positively \cite{zhou2023facilitating, li2024dq, cheng2022improving}, and have demonstrated its strength in the ESC task. Zhou et al. \cite{zhou2023facilitating} introduce a hybrid expert system targeting dialogue consistency into ESC to facilitate emotion change. Li et al. \cite{li2024dq} employ a heterogeneous graph construction and a Deep Q-learning algorithm for improved prediction of response strategies. Similarly, to select strategies with better results in the long-term horizon, Cheng et al. \cite{cheng2022improving} propose a heuristic algorithm to compute user feedback after the selection of specific strategies. Recent methods have increasingly focused on the effect of persona in dialogue systems. For instance, Cheng et al. \cite{cheng2023pal} dynamically extract persona information and model a strategy-based decoding strategy. Despite the focus on affective understanding and strategy prediction, there is a gap in addressing help-seeker's concealed cognitive states. Our work fills this gap by synergizing Dynamic Demonstration Retrieval with Cognitive-Aspect Situation Understanding to navigate the delicate balance between providing supportive responses and actionable suggestions in ESC.

\subsection{Retrieval-enhanced Response Generation}

Generative models, while show commendable performance in dialogue systems \cite{guo2018dialog}, can sometimes produce responses that are perceived as relatively mundane \cite{li2016diversity} (e.g., “I am not sure”) or suffer from hallucinations \cite{chen2022dialogved}. Retrieval-enhanced models address this by using reference utterances to guide response generation \cite{guu2018generating, weston2018retrieve, hofstatter2023fid}. For instance, retrieve reference utterances and utilize multi-level attention mechanisms (sentence-level and character-level) to mitigate hallucinations \cite{weston2018retrieve, shuster2021retrieval}. Moreover, additional research concentrate on constructing a skeleton and editing a selected retrieval reply to diminish the context and semantic gap between the retrieval context and the current context \cite{cai2019skeleton, cai2019retrieval, gupta2021controlling}. Techniques such as retrieval-rewriting further align retrieved responses with the current dialogue \cite{wu2019response}. For a more accurate personalized information, Zeng et al. \cite{zeng2023personalized} incorporating user preferences by developing a special attentive network. Maintaining consistent roles is crucial for constructing human-like dialogue systems, COSPLAY \cite{xu2022cosplay} enhances personality through a concept set framework. In spite of their success in general dialogue systems, applying this approach to the ESC domain remains underexplored. Our work pioneers the integration of retrieval-based methods within the ESC framework, dynamically selecting demonstration pairs and augmenting emotional support, presenting a novel paradigm that bridges the retrieval and cognitive understanding gap in support conversations.

\section{Our Approach}

\subsection{Task Formulation}
In this paper, we aim to address multi-turn Emotional Support Conversation task. In the setting of ESC, help-seekers consult with supporters with the anticipation of receiving effective emotional support. Through nuanced communication, supporters adeptly choose appropriate response strategies to aid help-seekers in overcoming challenges. In the $t$-th turn of the dialogue, given the dialogue context $c_t=(u_1, u_2, ..., u_{t-1})$, a set of $t-1$ seeker-supporter pairs, and the seeker's current post $p_t=(p^t_1,p^t_2,...,p^t_m)$ that consists of $m$ tokens, our goal is to generate the supportive response $r_t=r=(r^t_1, r^t_2, ..., r^t_n)$, comprising $n$ tokens. For convenience, the superscript $t$ associated with the $t$-th turn input is omitted in the following description.

\subsection{Overview of \ourwork}
As shown in Figure \ref{Fig:RECU}, \ourwork consists of three main components: \ding{172}\textit{Dynamic Demonstration Selector}, \ding{173}\textit{Cognitive-Aspect Situation Understanding}, and \ding{174}\textit{Multi-Knowledge Fusion Decoder}. 
First, the current post $p$ from user is fed into the \textit{Dynamic Demonstration Selector} to obtain candidate query-passage pairs. At the same time, the pre-trained COMET \cite{bosselut2019comet}, a generative commonsense reasoning model trained on ATOMIC \cite{sap2019atomic}, is utilized to generate the help-seeker's cognitive states. 
Second, the \textit{Cognitive-Aspect Situation Understanding} module is employed to refine cognitive states sequentially through its sub-modules, \textit{Cognitive Encoder}, \textit{Cognitive Refiner}, and \textit{Cognitive Selector}, thereby enhancing the quality of cognitive representations. 
Finally, the refined cognitive states and candidate query-passage pairs are forwarded to the cross-attention module, in conjunction with the dialogue history, facilitating the operation of the \textit{Multi-Knowledge Fusion Decoder}.

\subsection{Dynamic Demonstration Selector}
\begin{figure}[!t]
    \centering
    \includegraphics[width=0.95\linewidth]{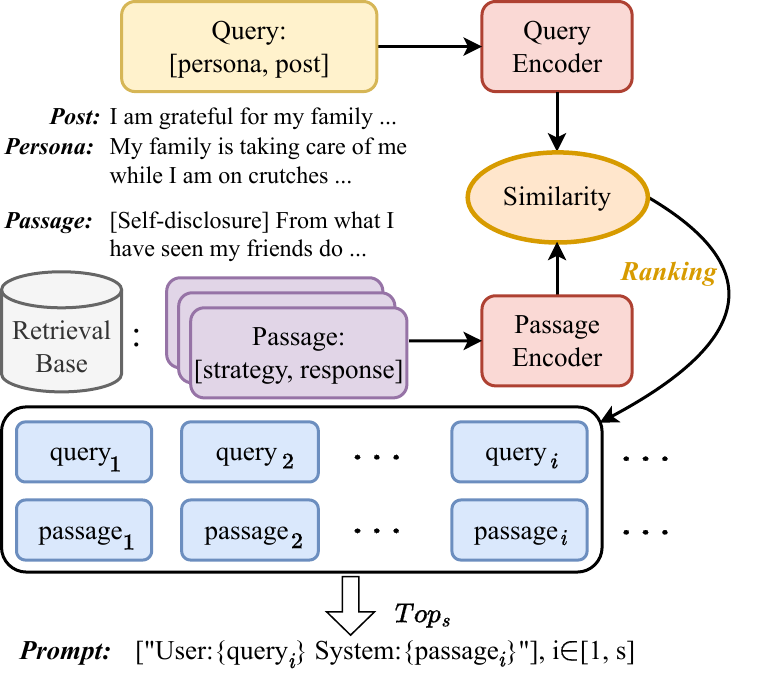}
    \caption{The \textit{Dynamic Demonstration Selector} architecture. The retrieval component (top) utilizes a pre-trained DPR to identify relevant responses within the training set, while the demonstration component (down) compiles the most pertinent user-system pairs, informed by the top $s$ results.}
    \label{Fig:DDS-module}
\end{figure}

In Emotional Support Conversation (ESC), contextually relevant and emotionally resonant responses are crucial. To achieve these, we introduce a \textit{Dynamic Demonstration Selector} that dynamically retrieves query-passage pairs that are semantically aligned with the current help-seeker's context and personal attributes. The selection process is visually represented in Figure \ref{Fig:DDS-module}.

The retrieval base plays a crucial role in personalized responses and can be compliantly collected in advance, such as various users' history dialogues.
In this work, we extract the strategic text $s$ and responsive text $r$ within each dialog turn from the training set of ESConv \cite{liu2021towards}, and compose them into a retrieval base as a set of passages $pas = [s, r]$. For example, as shown in Figure \ref{Fig:DDS-module}, $s$ is ``[Self-disclosure]'' and $r$ is `` From what I have seen my friends do ...''.

For the retrieval process, we leverage Dense Passage Retrieval (DPR) \cite{kwiatkowski2019natural, rajapakse2023dense} to facilitate the retrieval precision.
Specifically, a text passage encoder $E_P(\cdot)$ maps these passages into high-dimensional vectors for efficient retrieval. Simultaneously, the current help-seeker's post $p$ and associated persona $per$ are concatenated to form an input query $q = [p, per]$. 
We extract the help-seeker's persona field from the training set of PESConv \cite{cheng2023pal}, such as ``My family is taking care of me
while I am on crutches ...'' shown in Figure \ref{Fig:DDS-module}.
The query $q$ is then transformed into a vector representation using a query encoder $E_Q(\cdot)$:

\begin{equation}
    \text{sim}(q, pas) = E_Q(q)^{\top} E_P(pas).
\end{equation}

For the encoders $E_P$ and $E_Q$, we adopt ones of a DPR model from HuggingFace \footnote{https://huggingface.co/facebook/dpr-reader-single-nq-base}, which is pre-trained on real-world queries and excels in open-domain question answering.
By computing the vector similarity, the top $s$ passages with the highest degree of relevance are shortlisted as candidate responses. Corresponding queries $q_i$ for each retrieved strategic response $pas_i$ are also extracted. These elements are then merged to assemble a set of demonstrations $\mathcal{D}=\{d_1, d_2, ..., d_k\}$, with each demonstration defined as:

\begin{equation}
    d_i = \text{User:} +\ q_i + \text{System:} +\ pas_i,
\end{equation}
where $+$ indicates the textual join by separator with a blank.
The resulting demonstration pairs undergo further encoding to acquire their hidden representations:

\begin{equation}
    D_{prs} = d_1 \oplus d_2 \oplus \ldots \oplus d_s,
\end{equation}
\begin{equation}
    H_P = \text{Encoder}(D_{prs}),
\end{equation}
where ``$\oplus$'' denotes concatenation and $H_P$ encapsulates the encoded knowledge, with the total length capped at 512 tokens to ensure model efficiency.

The rationale behind the DDS is to extract and infuse the conversational model with nuanced, situation-appropriate responses, thereby elevating the emotional and cognitive relevance of the system's outputs. This approach is in stark contrast to static responses, granting our system the dynamism required to address the complex and varied emotional landscapes present in ESC.

\subsection{Cognitive-Aspect Situation Understanding}

Understanding the cognitive aspect of a help-seeker's situation is imperative for providing nuanced emotional support. To this end, our methodology integrates a \textit{Cognitive-Aspect Situation Understanding} module that distills the cognitive states of the dialogue participants from the context of their conversation.

The contextual dialogue history is aggregated into a single sequence, $CTX = u_1 \oplus u_2 \oplus \ldots \oplus u_N$, where $\oplus$ indicates concatenation operation, and $N$ is the number of utterances. This sequence is then encoded using a dedicated encoder:

\begin{equation}
    H_{CTX} = \text{Encoder}(CTX).
\end{equation}

Cognitive state can well describe the user's situation \cite{sabour2022cem}. To capture the cognitive states, we focus on four key situation-related aspects: want, need, intent, and effect. Each aspect is represented by tokens that are subsequently concatenated to the current post $p$. For each cognitive relation $r$, we invoke COMET to predict the corresponding cognitive states, $[com^r_1, com^r_2, \ldots, com^r_5]$. These predictions are amalgamated into a cognitive state sequence, $COM^r = com^r_1 \oplus com^r_2 \oplus \ldots \oplus com^r_5$, and encoded to obtain:

\begin{equation}
    E^r_C = \text{Encoder}(COM^r),
\end{equation}
with $E^r_C \in \mathbb{R}^{l_r \times d}$, where $l_r$ is the maximum sequence length for the cognitive states, and $d$ is the dimensionality of the hidden layer.

Our cognitive understanding module is comprised of three components: the \textit{Cognitive Encoder}, \textit{Cognitive Refiner}, and \textit{Cognitive Selector}. The representations obtained from the knowledge encoder, $E_C = E^{intent}_C \oplus E^{need}_C \oplus E^{effect}_C \oplus E^{want}_C$, pass through the \textit{Cognitive Encoder}. The hidden state corresponding to the $[CLS]$ token encapsulates the composite cognitive representation:

\begin{equation}
    H^{enc}C = \text{Enc}_{\text{enc}}(E_C),
\end{equation}
\begin{equation}
    h^{enc}_C = H^{enc}_C[0],
\end{equation}
where $H^{enc}_C \in \mathbb{R}^{l \times d}$, and $l$ is determined by the longer of the demonstration pairs or cognitive sequences.

To enrich the dialogue context with cognitive insights, we merge $H_{CTX}$ with $h^{enc}_C$ at the token level, allowing for a more granular integration of knowledge. This enriched representation is refined through the \textit{Cognitive Refiner} to generate cognition-enhanced context embeddings:

\begin{equation}
    U^{enc}_C = H_{CTX} \oplus h^{enc}_C,
\end{equation}
\begin{equation}
    H^{Ref}_C = \text{Enc}_{\text{ref}}(U^{enc}_C),
\end{equation}
where $H^{Ref}_C \in \mathbb{R}^{l \times 2d}$.

Our approach employs a sigmoid function to compute the importance score for each token within $H^{Ref}_C$. This score is then used to weight the tokens, emphasizing salient features within the cognition-refined context. A MLP, using ReLU activation, subsequently abstracts these weighted representations:

\begin{equation}
    H^{sel}_C = \text{Sigmoid}(H^{Ref}_C) \odot H^{Ref}_C,
\end{equation}
\begin{equation}
    H_C = \text{MLP}(H^{sel}_C),
\end{equation}
which results in $H_C \in \mathbb{R}^{l \times d}$. Through this process, our model acquires a deeper, more abstracted meaningful understanding of the cognitive states that underpin the help-seeker's narrative, enabling the generation of responses that are not only empathetic but also cognitively aligned with the seeker’s expressed needs and desires.

\subsection{Multi-Knowledge Fusion Decoder}

To fully capitalize on the two distinct knowledge sources retrieved demonstration pairs and cognitive states — our approach introduces a \textit{Multi-Knowledge Fusion Decoder}. This component ensures the coherent integration of context-sensitive knowledge into the response generation process, which is pivotal for crafting supportive conversations that are both relevant and cognitively congruent.

The decoder's mechanism begins with a dual cross-attention operation that separately assimilates the encoded demonstration pairs and cognitive states with the dialogue history. This process establishes a bidirectional flow of information and aligns the knowledge representations with the context of the conversation. We compute the affinity scores between the historical context ($H_{CTX}$) and the encoded knowledge sources ($H_P$ for demonstration pairs and $H_C$ for cognitive states), yielding updated context representations:

\begin{equation}
    Z^{P}_D = \text{Softmax}(H_{CTX} \cdot H^{\top}_P) \cdot H_P,
\end{equation}
\begin{equation}
    Z^{C}_D = \text{Softmax}(H_{CTX} \cdot H^{\top}_C) \cdot H_C,
\end{equation}
\begin{equation}
    {\tilde{H}^{P}_D} = \text{LayerNorm}(H_{CTX} + Z^{P}_D),
\end{equation}
\begin{equation}
    {\tilde{H}^{C}_D} = \text{LayerNorm}(H_{CTX} + Z^{C}_D).
\end{equation}

Similarly, to align the historical context with the two knowledge entities, analogous computations are performed to derive contextually relevant demonstration and cognitive state representations $\tilde{H}^D_P$ and $\tilde{H}^D_{C}$.

To balance and integrate these aligned features, we employ a weighted aggregation strategy, resulting in a composite hidden state vector, $H_{fin}$, that serves as the input for the decoder:

\begin{equation}
    H_{fin} = \lambda_1 \cdot H_D + \lambda_2 \cdot \tilde{H}^P_D + \lambda_3 \cdot \tilde{H}^C_D + \lambda_4 \cdot \tilde{H}^D_P + \lambda_5 \cdot \tilde{H}^D_C,
\end{equation}
\begin{equation}
    \lambda_i = \frac{e^{w_i}}{\sum_j e^{w_j}},
\end{equation}
where $i, j \in \{1, 2, 3, 4, 5\}$. The adaptive parameters $w$ are initialized equally but can be optimized during training to reflect the salience of each feature.

\begin{algorithm}[t]
  \caption{Algorithm for \ourwork.}
  \label{Alg:flow}
  \KwData{$\{(p_t, c_t, r_t\}^{N_{data}}_{n=1}$, a dataset of multi-turn dialogues.}
  \KwData{Hyperparameters: $N_{epochs}\in\mathbb{N}$.}
  \KwResult{$\hat{\theta}$, the trained parameters.}

  \SetAlgoNlRelativeSize{0}
  \SetAlgoNlRelativeSize{-1}
  \SetAlgoNlRelativeSize{1}
  \SetAlgoNlRelativeSize{0}

  \For{$i \leftarrow 1$ \KwTo $N_{epochs}$}{
    \For{$n \leftarrow 1$ \KwTo $N_{data}$}{
      \For{$t \leftarrow 1$ \KwTo $N_{turns}$}{
        \tcp{\ding{172} Dynamic Demonstration Selector}
        $Pairs \leftarrow DDS(p_t)$ \\
        $H_{P} \leftarrow \text{Encoder}(Pairs)$ \\

        \tcp{\ding{173} Cognitive Understanding}
        $Commonsense \leftarrow COMET(p_t)$ \\
        $H_{CTX} \leftarrow \text{Encoder}(c_t)$ \\
        $E_{C} \leftarrow \text{Encoder}(Commonsense)$ \\
        $H^{enc}_C \leftarrow \text{Enc}_\text{enc}(H_C)$ \\
        $H^{Ref}_{C} \leftarrow \text{Enc}_\text{ref}(H^{enc}_C)$ \\
        $H_{C} \leftarrow \text{Enc}_\text{cog}(H^{Ref}_{C})$ \\

        \tcp{\ding{174} Multi-Knowledge Fusion Decoder}
        $\tilde{H}^P_D, \tilde{H}^D_P \leftarrow \text{CoAttn}(H_{P}, H_{CTX})$ \\
        $\tilde{H}^C_D, \tilde{H}^D_C \leftarrow \text{CoAttn}(H_{C}, H_{CTX})$ \\
        $H_{fin} \leftarrow \lambda_1{\cdot}H_{CTX}+\lambda_2{\cdot}\tilde{H}^P_D+\lambda_3{\cdot}\tilde{H}^C_D+\lambda_4{\cdot}\tilde{H}^D_P+\lambda_5{\cdot}\tilde{H}^D_C$ \\
        $\hat{H}_{fin}=\text{LayerNorm}(H_{fin})$ \\

        \tcp{Gradient Optimization}
        $P(\theta) \leftarrow \text{Decoder}(\hat{H_{fin}}|\theta)$ \\
        $loss(\theta) = -\sum^{l-1}_{t=1}{\log}P(\theta)$ \\
        $\theta \leftarrow \theta - \eta{\cdot}\nabla\text{loss}(\theta)$
      }
    }
  }
  
  \Return{$\hat{\theta} = \theta$}
  
\end{algorithm}

To promote uniformity across the feature dimensions and enhance model generalization, we standardize each feature of $H_{fin}$ using layer normalization:

\begin{equation}
    \hat{H}_{fin} = \text{LayerNorm}(H_{fin}).
\end{equation}

The generation of the target response $r = (r_1, r_2, \ldots, r_n)$, a sequence of length $n$, is predicated upon the fused knowledge-enriched hidden state:

\begin{equation}
    P(r_t | r_{<t}, C) = \text{Decoder}(E_{r_{<t}}, \hat{H})_{fin},
\end{equation}
where $E$ denotes the embedding of the response generated thus far, and $r_{<t}$ represents the tokens generated up to the current timestep. Our training objective is defined by the negative log-likelihood of the correct response:

\begin{equation}
    \mathcal{L} = -\frac{1}{n} \sum^n_{t=1} \text{log} P(r_t | C, r_{<t}).
\end{equation}

We summarize the pseudocode of the proposed \ourwork in Algorithm \ref{Alg:flow}.
Through these modules, we aim to synthesize a comprehensive response that not only demonstrates supportive but also reflects a deep cognitive understanding for ESC systems.

\section{Experimental Setting}

\subsection{Dataset}
For a fair evaluation, we examine our approach on the Emotional Support Conversation Dataset (ESConv), a high-quality crowdsourced dataset that is constructed from interactions between help-seekers and supporters and widely used by SOTA methods in ESC \cite{tu2022misc, peng2022control, deng2023knowledge, peng2023fado, zhao2023transesc, cheng2023pal}. 
There are 1,300 lengthy conversations in ESConv, and the average number of turns per conversation is 29.8, which has more interactions and turns than empathic dialogue \cite{rashkin2019towards}. 
In each turn of dialogue, supporters provide responses, with an average length of 20.2 turns, guided by specific response strategies, offering eight choices, including Question, Restatement or Paraphrasing, Reflection of Feelings, Self-disclosure, Affirmation and Reassurance, Providing Suggestions, Information, and Others. 
Following Liu et al. \cite{liu2021towards}, we also divide the dataset into training (80\%), validation (10\%), and test (10\%) sets. 

\subsection{Metrics}
For a comprehensive evaluation, we adopt nine metrics used in existing works \cite{peng2023fado, peng2022control}: ACC, PPL, B-1, B-2, B-3, B-4, D-1, D-2 and R-L. 
The first metric ACC indicates the accuracy of predicted response strategies (one of the eight ground-truth choices in ESConv \cite{liu2021towards}), while the rest of the metrics are conventional measures for evaluating response generation.
Furthermore, we employ $\bar{s}_{\text{norm}}$ to impartially and comprehensively assess the overall performance. Now, let's introduce these evaluation metrics below.

\begin{table*}[!t]
  \caption{Results of automatic evaluation on ESConv, where the values of baselines are sourced from public papers. The bolded and \underline{underlined} results indicate best and second-best result respectively. \textbf{$\bar{s}_{norm}$} is normalized average score over all metrics.}
  \label{Tab:results}
  \begin{tabular}{lcccccccccc}
    \toprule
    \textbf{Model} & \textbf{ACC$\uparrow$} & \textbf{PPL$\downarrow$} & \textbf{B-1$\uparrow$} & \textbf{B-2$\uparrow$} & \textbf{B-3$\uparrow$} & \textbf{B-4$\uparrow$} & \textbf{D-1$\uparrow$} & \textbf{D-2$\uparrow$} & \textbf{R-L$\uparrow$} & \textbf{$\bar{s}_{norm}\uparrow$}\\
    \midrule
    \textbf{MIME} (2020 EMNLP) \cite{majumder2020mime} & - & 43.27 & 16.15 & 4.82 & 1.79 & 1.03 & 2.56 & 12.33 & 14.83 & 0.00 \\
    \textbf{BlenderBot-Joint} (2021 ACL) \cite{liu2021towards} & 27.72 & 18.11 & 16.99 & 6.18 & 2.95 & 1.66 & 3.27 & 20.87 & 15.13 & 0.35 \\
    \textbf{MISC} (2022 ACL) \cite{tu2022misc} & 31.34 & 16.28 & - & 6.60 & - & 1.99 & 4.53 & 19.75 & 17.21 & 0.56 \\
    \textbf{GLHG} (2022 IJCAI) \cite{peng2022control} & - & 15.67 & 19.66 & 7.57 & 3.74 & 2.13 & - & - & 16.37 & 0.64 \\
    \textbf{KEMI} (2023 ACL) \cite{deng2023knowledge} & - & 15.92 & - & 8.31 & - & 2.51 & - & - & 17.05 & 0.78 \\
    \textbf{FADO} (2023 KBS) \cite{peng2023fado} & 32.41 & 15.52 & - & 8.31 & 4.00 & 2.66 & 3.80 & 21.39 & 18.09 & 0.71 \\
    \textbf{SUPPORTER} (2023 ACL) \cite{zhou2023facilitating} & - & 15.37 & 19.50 & 7.49 & 3.58 & - & 4.93 & 27.73 & - & 0.72 \\
    \textbf{TransESC} (2023 ACL) \cite{zhao2023transesc} & 34.71 & 15.85 & 17.92 & 7.64 & 4.01 & 2.43 & 4.73 & 20.48 & 17.51 & 0.74 \\
    \textbf{PAL} (2023 ACL) \cite{cheng2023pal} & 34.51 & 15.92 & - & 8.75 & - & 2.66 & \underline{5.00} & \underline{30.27} & 18.06 & 0.87 \\
    \midrule
    \textbf{\ourwork} (ours)& \textbf{35.32} & \textbf{15.43} & \underline{21.20} & \underline{9.01} & \underline{4.81} & \underline{2.94} & 4.97 & 26.21 & \textbf{18.71} & \underline{0.92} \\
    \textbf{\ourwork w/ norm} (ours) & \underline{34.87} & \underline{15.47} & \textbf{21.91} & \textbf{9.25} & \textbf{4.89} & \textbf{2.98} & \textbf{5.98} & \textbf{30.55 }& \underline{18.22} & \textbf{0.98} \\
    \bottomrule
  \end{tabular}
\end{table*}

\textbf{Perplexity (PPL)}:  An evaluation metric that quantifies how well a model predicts sequences of words, with lower values indicating more accurate and confident predictions.

\textbf{BLEU-n (B-n)} \cite{papineni2002bleu}: It is commonly used for automatic evaluation of machine translation results, but also for evaluating other natural language generation tasks, where $n\in\{1, 2, 3, 4\}$. Higher B-n scores suggest a better match between the system generated response and human references.

\textbf{DISTINCT-n (D-n)} \cite{li2016diversity}: This metric is often used to evaluate the diversity of generated text by calculating the proportion of unique words in relation to the total number of words, where $n\in\{1, 2\}$. Higher D-n scores indicate more diverse and varied language generation.

\textbf{ROUGE} \cite{lin2004rouge}: It is a set of metrics that measure the overlap between automatically generated text and reference text. \textbf{ROUGE-L (R-L)} is one of them, focusing on the Longest Common Subsequence. R-L is commonly used for evaluating the quality of text summarization systems, with higher scores indicating better summary generation.

$\bar{s}_{\text{norm}}$: For a fair and comprehensive comparison, we use an aggregate measure of performance across all considered metrics.
Specifically, we first normalize the score for each evaluation metric as $s_{\text{norm}} = (s - s_{\text{worst}})/(s_{\text{best}} - s_{\text{worst}})$, where $s$ stands for the raw score obtained for the metric, $s_{\text{worst}}$ represents the worst score observed across all methods being compared for that metric, and $s_{\text{best}}$ signifies the best score for the same metric. Then we synthesizes these individual normalized scores into a single metric by computing their arithmetic mean and refer it as $\bar{s}_{\text{norm}}$.

\subsection{Training Details}
For a fair comparison, we use the backbone of our model as the same as the setup in \cite{liu2021towards}, which is based on the chat conversation model Blenderbot-small \cite{roller2021recipes}. 
The maximum input length for the encoder is 512 and for the decoder is 50. For both the retrieved demonstration pair sequence and the generated cognitive state sequence, the maximum sequence length input into the encoder is set at 512. Model training is conducted on a single 3090Ti GPU, initializing the learning rate at 1.5e-5. The batch sizes of the training set and validation set are set to 4 and 16, respectively.  The optimizer employed is Adam with parameters beta1=0.9 and beta2=0.999. The model is iteratively trained for up to 10 epochs, and the checkpoint with the lowest ppl on the validation set is selected after 6 epochs. Following Cheng et al. \cite{cheng2023pal}, during the inference stage, the decoding algorithm employs sampling with $top_k$=10, $top_p$=0.9, and a repetition penalty of 1.03. 

\subsection{Baselines}

To examine the effectiveness of our model, we compare \ourwork with 8 SOTA baselines.

\textbf{MIME} \cite{majumder2020mime}: A Transformer-based model that utilizes multiple decoders for different emotions to imitate the speaker's emotions.

\textbf{BlenderBot-Joint} \cite{liu2021towards}: An open-domain dialogue model has the ability to respond empathically, which is trained with multiple communication skills. On this basis, the model is fine-tuned on the ESConv dataset and is combined with strategy prediction. 

\textbf{MISC} \cite{tu2022misc}: It is a seq2seq model that utilizes commonsense to obtain fine-grained emotion information and uses a mixed-strategy to generate responses.

\textbf{GLHG} \cite{peng2022control}: The global-local hierarchical graph network models the relationship between global causes and local intentions.

\textbf{KEMI} \cite{deng2023knowledge}: It is a mix-initiative framework that first performs query expansion followed by subgraph retrieval on the knowledge graph, giving both help-seekers and supporters the opportunity to lead the dialogue.

\textbf{FADO} \cite{peng2023fado}: A dual-level strategy selection mechanism is based on turn and conversation, and selects user-related strategies through feedback from help-seekers.

\textbf{SUPPORTER} \cite{zhou2023facilitating}: A reinforcement learning model SUPPORTER based on a hybrid expert model is proposed, and positive emotion stimulation and dialogue coherence rewards are designed to guide policy learning.

\textbf{TransESC} \cite{zhao2023transesc}: It considers three fine-grained transitions (semantic, strategy and emotional) between each conversation turn and build a state transition graph.

\textbf{PAL} \cite{cheng2023pal}: It dynamically extracts persona information and models a strategy-based decoding strategy.

\section{Results and Analysis}
\subsection{Automatic Evaluation}

To demonstrate the effectiveness of our approach, we conduct a comprehensive comparison with numerous SOTA models, and the results for each method are presented in Table \ref{Tab:results}. Our approach achieves a substantial overall performance improvement ($\bar{s}_{norm}$) of up to \textbf{13.79\%}, due to the effective of dynamic in-context learning and cognitive state understanding. Significant improvement over the strongest baseline is observed in strategy prediction accuracy (ACC), with the accuracy reaching \textbf{35.32\%}. Our B-n is notably higher, with a \textbf{12.03\%} enhancement in the most challenging B-4, indicating a higher degree of alignment between our generated responses and the reference responses. Similarly, achieving \textbf{18.71} on R-L, our responses demonstrate superior effectiveness. In terms of response diversity (D-n), our method achieves competitive performance, with a \textbf{19.60\%} augmentation in D-1.

Comparing to the model focusing on empathetic responses (MIME), \ourwork provides positive responses to address help-seeker's current situation, resulting in a \textbf{22.86\%} improvement in R-L. 
When compared with models exploiting commonsense knowledge extracted by Comet (e.g., MISC, GLHG, KEMI, TransESC), our method explicitly models the cognitive understanding of the interlocutor's situation and retrieves the most similar query-passage pairs, achieving the $\bar{s}_{\text{norm}}$ score of \textbf{0.92}. 
In addition, we design another training strategy that leverages normalization operations during training to make the model converge more easily during training and helps to alleviate the problems of vanishing and exploding gradients. 

\begin{figure}[!t]
  \centering
  \includegraphics[width=0.9\linewidth]{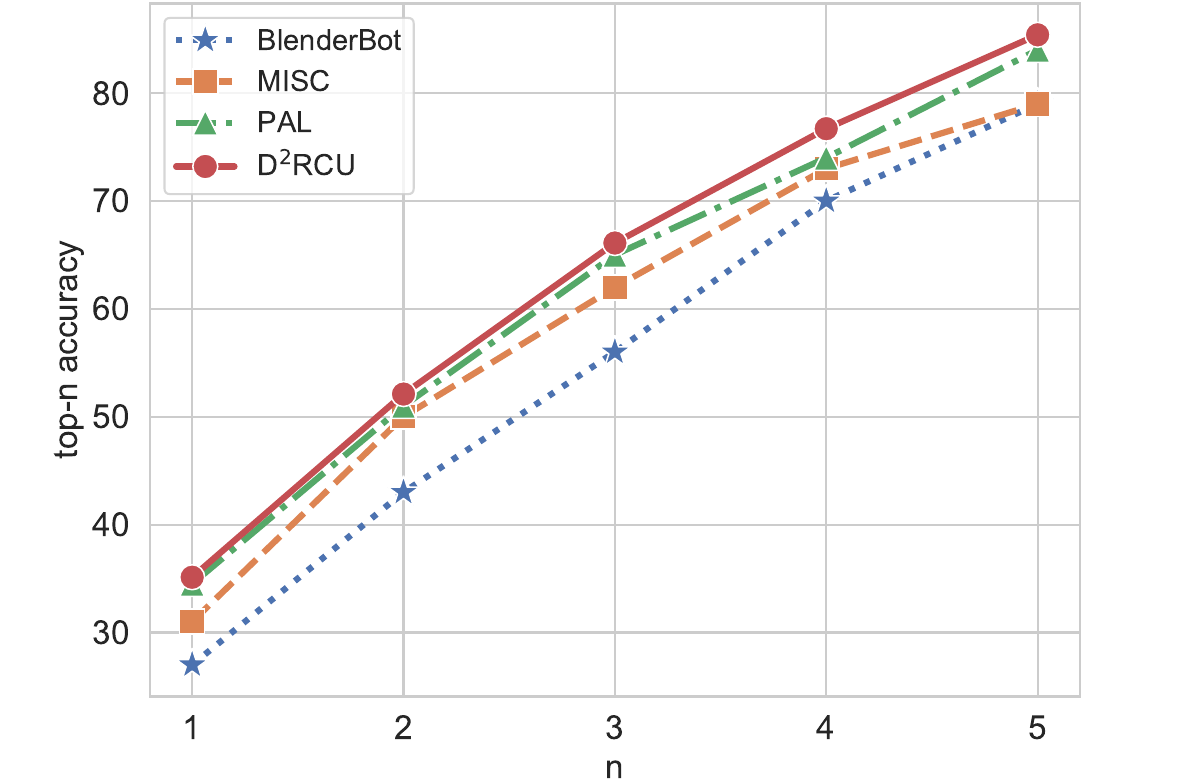}
  \caption{The top-n accuracy of predicted strategies. \ourwork consistently gains best performance.}
  \label{Fig:ACC}
\end{figure}

In Figure \ref{Fig:ACC}, we compare the accuracy of strategy prediction across different precision levels, and the results show that our \ourwork achieves optimal performance across all precision levels. Compared to PAL, which also incorporates persona information, our model achieves an average improvement of \textbf{0.85\%}. Furthermore, compared to MISC, which similarly utilizes COMET for extracting commonsense knowledge, our model demonstrates more robust performance. This underscores the enhanced capabilities of our dynamic retrieval and cognitive understanding in comprehending the user's situation and delivering contextually appropriate responses.

\begin{table}
  \caption{The results of ablation study for \ourwork. COG and EMO indicates utilizing cognitive and affective knowledge respectively.}
  \label{Tab:ablation}
  \small
  \begin{tabular}{lccccccc}
    \toprule
    \textbf{Model} & \textbf{ACC$\uparrow$} & \textbf{PPL$\downarrow$} & \textbf{B-1$\uparrow$} & \textbf{B-2$\uparrow$} & \textbf{B-3$\uparrow$} & \textbf{B-4$\uparrow$} & \textbf{R-L$\uparrow$} \\
    \midrule
    \textbf{\ourwork} & \textbf{35.32} & \textbf{15.43} & \textbf{21.20} & \textbf{9.01} & \textbf{4.81} & \textbf{2.94} & \textbf{18.71} \\
    \midrule
    w/o DDS & 33.95 & 15.91 & 20.12 & 8.17 & 4.21 & 2.50 & 18.25 \\
    w/o Pairs & 31.31 & 16.17 & 19.79 & 8.03 & 4.07 & 2.37 & 17.99 \\
    \midrule
    w/o COMET & 35.38 & 15.46 & 19.54 & 8.26 & 4.44 & 2.74 & 18.51 \\
    w/o CU & 32.91 & 15.79 & 20.23 & 8.15 & 4.09 & 2.31 & 17.69 \\
    \midrule
    COG+EMO & 35.05 & 15.43 & 21.15 & 8.91 & 4.73 & 2.88 & 18.64 \\
    EMO & 34.13 & 15.54 & 21.21 & 8.92 & 4.71 & 2.83 & 18.49 \\
    \bottomrule
  \end{tabular}
\end{table}

\subsection{Ablation Study}

In order to investigate the impact of each component within \ourwork, we conducted ablation studies on the \textit{Dynamic Demonstration Selector} (DDS, \textit{the blue box} in Figure \ref{Fig:RECU}) component and the \textit{Cognitive-Aspect Situation Understanding} (COMET, \textit{the yellow box} in Figure \ref{Fig:RECU}) component. The results are summarized in Table \ref{Tab:ablation}. At the same time, we explore the influence of cognition and affectiveness on the ESC task. 

Firstly, we remove the entire DDS component (w/o DDS), resulting in a performance reduction compared to the original method, especially with B-4 decreasing by \textbf{14.97\%}. To assess the effectiveness of in-context learning, we concatenate responses from retrieved responses without using query-passage pairs (w/o PAIRS). The results indicate that without using retrieval pairs, i.e., without employing the in-context method, B-4 would further decrease by \textbf{5.2\%}, possibly due to the noise outweighing the useful information from the retrieved responses. 
Secondly, we remove the entire COMET part (w/o COMET), leading to a decrease in performance, although not as pronounced as when removing DDS, indicating the higher importance of the DDS. We exclude the Cognitive-Aspect Situation Understanding component (w/o CU) to evaluate its importance. It is found that without CU, cognitive states are not understood, resulting in a decrease of \textbf{21.43\%} in B-4. Finally, while \ourwork fully leverages the cognitive knowledge extracted from COMET, we compare it with models that utilize both cognitive and affective knowledge (COG+EMO) and models that only utilize affective knowledge (EMO). The observations indicate that the model incorporating cognitive understanding outperforms the model relying solely on the affective aspect. Collectively, above results verify the effectivness and necessity of the components of \ourwork.

\subsection{Human Evaluation}

\begin{table}
  \caption{Human evaluation results (\ourwork against others).}
  \label{Tab:human}
  \begin{tabular}{lcccc}
    \toprule
    \multirow{2}{*}{Competitors with \ourwork} & \multicolumn{3}{c}{PAL} \\
    \cmidrule(lr){2-4}
    & Win (\%) & Lose (\%) & Tie (\%) \\
    \midrule
    Fluency & \textbf{44} & 30 & 26 \\
    Coherence & \textbf{44} & 28 & 28 \\
    Comforting & \textbf{46} & 30 & 24 \\
    Suggestion & \textbf{44} & 32 & 24 \\
    Informativeness & \textbf{34} & 34 & 32 \\
    \midrule
    Overall & \textbf{52} & 32 & 16 \\
    \bottomrule
  \end{tabular}
\end{table}

\begin{figure*}[!ht]
    \centering
    \includegraphics[width=\linewidth]{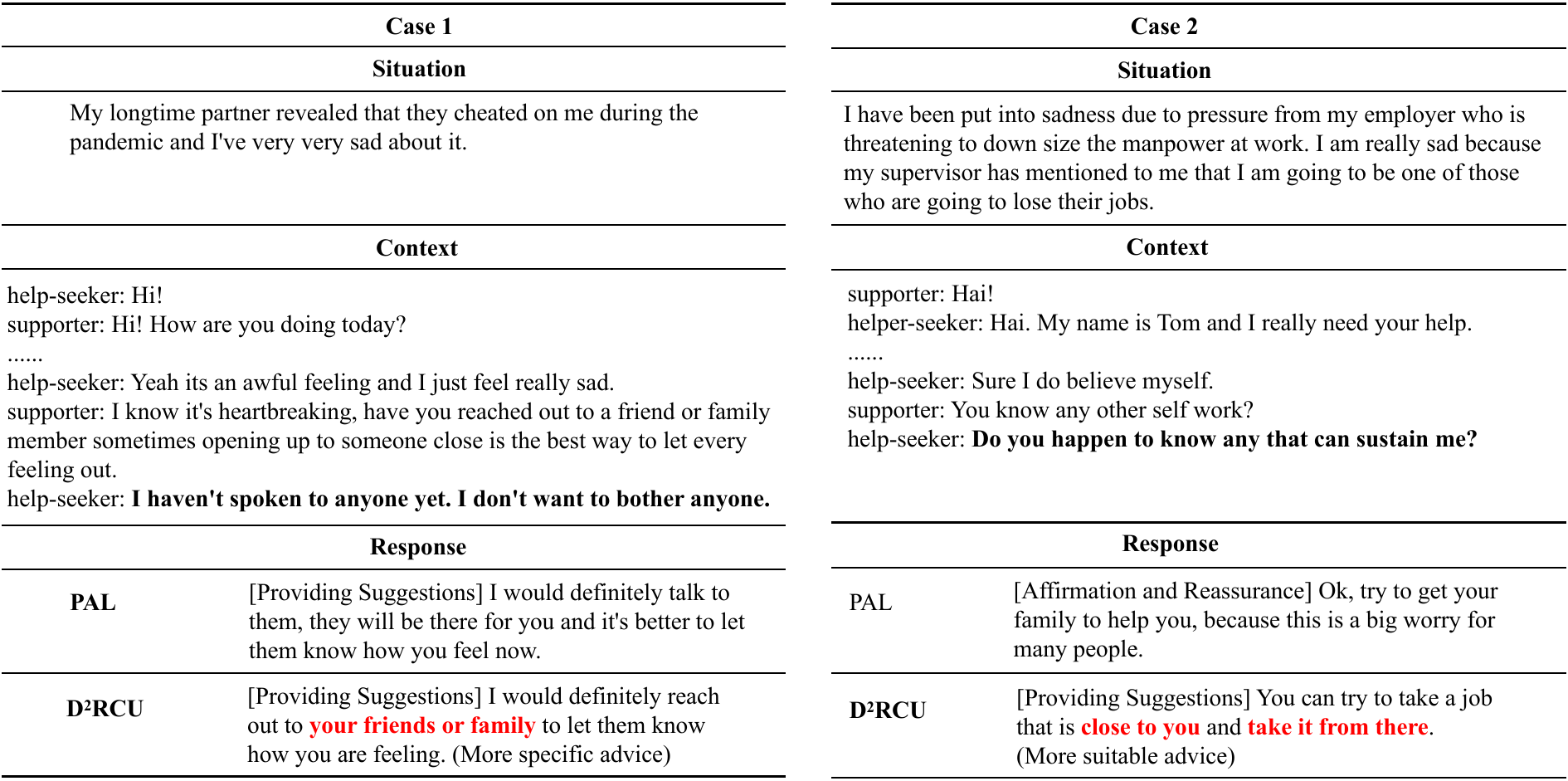}
    \caption{Comparative results from our case studies of \ourwork and state-of-the-art PAL.}
    \label{Fig:Case 3}
\end{figure*}

We randomly select 100 sample dialogues from the test set for human evaluation. Subsequently, we recruited four volunteers through crowdsourcing, all of whom are graduate students in the field of natural language processing and possess a background in dialogue systems, to compare our \ourwork with the current state-of-the-art model PAL. Following the same evaluations with prior studies \cite{liu2021towards, cheng2023pal}, the volunteers assess the quality of the generated responses based on six criteria: (1) Fluency: The degree to which responses are fluent and understandable; (2) Coherence: How well responses align with the context of the conversation; (3) Comforting: The ability of responses to provide comfort to the person seeking help; (4) Suggestion: How well responses offer advice or suggestions to the person seeking help; ( 5) Informativeness: The amount of information contained in the responses; (6) Overall: The overall effectiveness of the generated responses. Each volunteer provided ratings for both models across these criteria, contributing to a comprehensive evaluation of the dialogue generation quality. 

The results of human evaluation are presented in Table \ref{Tab:human}, which compares responses generated by our \ourwork with those produced by PAL. The outcomes show that \ourwork achieves significant improvements over the strongest baseline in supportiveness, particularly in terms of Fluency, Coherence, Comforting and Suggestion. The high scores in Fluency and Coherence indicate that during counseling, the help-seeker can engage in a smoother conversation with the supporter. Comforting and Suggestion are key metrics for evaluating the supportability of responses, and their excellent results show that \ourwork performs better in terms of supporting conversations and is more beneficial in providing comfort and advice to the person seeking help. Finally, we outperform PAL with higher win rates (\textbf{52\%} v.s. \textbf{32\%}) in overall metrics, indicating that our work can generate responses more suitable for emotional support conversations than other models. 

\subsection{Parameter Analysis}

\begin{table}[!t]
  \centering
  \caption{Analysis of the model performance based on the top-s demonstrations in the Dynamic Demonstration Selector.}
  \label{Tab:TopK} 
  \begin{tabular}{ccccccc}
    \toprule
    \textbf{Top s} & \textbf{ACC$\uparrow$} & \textbf{B-1$\uparrow$} & \textbf{B-2$\uparrow$} & \textbf{B-3$\uparrow$} & \textbf{B-4$\uparrow$} & \textbf{R-L$\uparrow$} \\
    \midrule
    $s = 1$ & 31.38 & 19.23 & 7.83 & 3.95 & 2.29 & 17.74 \\
    $\textbf{s = 3}$ & 35.32 & \textbf{21.21} & \textbf{9.01} & \textbf{4.83} & \textbf{2.94} & \textbf{18.71} \\
    $s = 5$ & 34.78 & 18.75 & 7.86 & 4.22 & 2.57 & 17.83 \\
    $s = 10$ & \textbf{35.38} & 19.10 & 8.00 & 2.63 & 2.63 & 18.03 \\
    \bottomrule
  \end{tabular}
\end{table}

In this section, we explore the impact of different numbers of retrieval demonstration pairs on the method's performance, and the results are presented in Table \ref{Tab:TopK}. From the table, it can be observed that as the number of retrieved demonstration pairs increases, the overall performance shows a trend of first increasing and then decreasing. The phenomenon is attributed to the fact that when there is only one sample pair, \ourwork lacks sufficient reference information. As the number of demonstration pairs reaches 5 or 10, a considerable amount of noise information is introduced due to retrieval based on relevance with the help-seeker's queries. While there is an improvement in ACC when top $s$ = 10, this enhancement comes at the cost of reduced scores in B-1/2/3/4 and R-L. After comprehensive consideration, we choose to retrieve the 3 most relevant demonstration pairs as the input of in-context learning, which yields a better overall result and is efficient to train.

\subsection{Case Study}

We analyze two case studies to compare the responses generated by \ourwork and the strongest baseline PAL. The results are visualized in Figure \ref{Fig:Case 3}, which demonstrate how each model provides emotional support conversation with distinct strategies.
In the first case study, we can observe that both models identify the strategy as \textit{``Providing Suggestions''}, which aligns with the conversational needs. 
However, \ourwork's retrieval-based augmentation component, coupled with the cognitive-based understanding component, allows it to generate a response that includes concrete advice such as \textit{``consider reaching out to friends or family for support,''} in contrast to PAL's more abstract suggestion to rely on \textit{``them''}. 
This comparison showcases \ourwork's ability to produce tailored and actionable guidance.
In the second case, the help-seeker inquires about resources for sustenance. In response to such a request, it is critical that the supporter provides comprehensive assistance. The response from \ourwork is constructed to give pertinent advice that is not only practical but also carries a positive and substantive influence. This illustrates \ourwork's proficiency in handling complex help-seeking inquiries with thoughtful and impactful responses.

\section{Conclusion}

This paper addresses pivotal challenges in Emotional Support Conversation (ESC) systems by presenting \ourwork, a novel approach that harmonizes dynamic demonstration retrieval with advanced cognitive understanding. We demonstrated how this methodology significantly enhances the empathetic and contextually relevant response generation of ESC systems, outperforming numerous state-of-the-art models with a marked 13.79\% improvement across key performance metrics. The introduction of an innovative retrieval mechanism and the incorporation of cognitive relationships from external knowledge source (ATOMIC) have proven to be instrumental in deepening the system's insight into the implicit mental states of help-seekers, facilitating responses with heightened empathy and cognitive awareness.

By improving the quality of supportive responses in ESC systems, we hope our work can inspire further research and applications that bring positive outcomes in mental health interventions and emotional counseling of users. 
Future research could optimize demonstration retrieval processes, introduce additional cognitive knowledge such as via foundation models \cite{jiao2024enhancing,chen2024datajuicer}, and adapt our system for a broader range of conversational scenarios.

\bibliographystyle{ACM-Reference-Format}
\balance
\bibliography{sample-base}

\end{document}